\documentclass[journal]{IEEEtran}
\usepackage{amsmath,amsfonts}
\usepackage{array}
\usepackage[caption=false,font=normalsize,labelfont=sf,textfont=sf]{subfig}
\usepackage{textcomp}
\usepackage{stfloats}
\usepackage{url}
\usepackage{verbatim}
\usepackage{graphicx}
\usepackage{cite}
\def\eg{\emph{e.g.}} 
\def\ie{\emph{i.e.}} 

\def\etc{\emph{etc.}}

\usepackage{graphicx}
\usepackage{mathptmx}      
\usepackage{times}
\usepackage{epsfig}
\usepackage{graphicx}
\usepackage{amsmath}
\usepackage{amssymb}
\usepackage{caption}
\usepackage[pagebackref=true,breaklinks=true,letterpaper=true,colorlinks,bookmarks=false]{hyperref}
\usepackage{enumerate}
\usepackage{booktabs} 
\usepackage{array} 
\usepackage{multirow}
\usepackage{colortbl}
\usepackage{animate}
\usepackage{bbding}
\usepackage{blindtext}
\usepackage[ruled,lined,linesnumbered]{algorithm2e}
\usepackage{color}

\definecolor{gray1}{RGB}{120,120,120}

\begin{document}
\title{VMA: Divide-and-Conquer \underline{V}ectorized \underline{M}ap \underline{A}nnotation System for Large-Scale Driving Scene}
\author{Shaoyu Chen, Yunchi Zhang, Bencheng Liao, Jiafeng Xie, Tianheng Cheng, Wei Sui, Qian Zhang, Chang Huang, Wenyu Liu,~\IEEEmembership{Senior Member,~IEEE}, and Xinggang Wang,~\IEEEmembership{Member,~IEEE}
\thanks{This work was in part supported by the National Key Research and Development Program of China (No. 2022YFB4500602) and NSFC (No. 62276108).}
\thanks{Shaoyu Chen, Yunchi Zhang, Bencheng Liao, Tianheng Cheng, Wenyu Liu, and Xinggang Wang are with the School of Electronic Information and Communications, Huazhong University of Science and Technology, Wuhan 430074, P.R. China (e-mail: \texttt{\{shaoyuchen, zyc10ud, bcliao, thch, liuwy, xgwang\}@hust.edu.cn}).}%
\thanks{Jiafeng Xie, Wei Sui, Qian Zhang, and Chang Huang are with Horizon Robotics (e-mail: \texttt{\{jiafeng.xie, wei.sui, qian01.zhang, chang.huang\}@horizon.ai}).}
\thanks{Shaoyu Chen and Yunchi Zhang contributes equally to this work. Xinggang Wang is the corresponding author.}
}

\markboth{}%
{Chen \MakeLowercase{\textit{et al.}}: Divide-and-Conquer Vectorized Map Annotation System for Large-Scale Driving Scene}

\maketitle

\begin{abstract}
     High-definition (HD) map serves as the essential infrastructure of autonomous driving. In this work, we build up a systematic vectorized map annotation framework (termed VMA) for efficiently generating HD map of large-scale driving scene. 
     We design a divide-and-conquer annotation scheme to solve the spatial extensibility problem of HD map generation, and abstract map elements with a variety of geometric patterns as unified point sequence representation, which can be extended to most map elements in the driving scene.
     VMA is highly efficient and extensible, requiring negligible human effort, and flexible in terms of spatial scale and element type.
     We quantitatively and qualitatively validate the annotation performance on  real-world urban and highway scenes, as well as NYC Planimetric Database.
     VMA can significantly improve  map generation efficiency and require little human effort. On average VMA takes 160min for annotating a scene with a range of hundreds of meters, and reduces 52.3\% of the human cost, showing great application value.
\end{abstract}

\begin{IEEEkeywords}
HD Map Generation, Divide-and-Conquer Scheme, Auto Labeling, Scene Reconstruction, Vectorized Representation, Autopilot, Scene Understanding.
\end{IEEEkeywords}

\section{Introduction}
\IEEEPARstart{H}{igh-definition} (HD) map contains a wide range of  static map elements and  encodes rich prior information of the driving scene,
serving as the essential infrastructure of autonomous driving.
However, map generation  requires laborious human effort and high expenses,
restricting the application and  deployment of autonomous driving systems. 
In this work, we propose a systematic Vectorized Map Annotation (VMA) framework to improve the map generation efficiency of large-scale driving scenes.

VMA is  highly extensible and flexible in terms of map element types.
HD map is composed of diverse map elements with a variety of geometric patterns:
line-shaped elements (road curb, lane divider,  stop line, \etc{}), regular-shaped elements (arrow, speed bump,  diamond marking, \etc{}), and closed-shaped regions (crosswalk, road intersection, diversion, \etc{}).
Though some image processing methods~\cite{canny,Zhang2021Research} and segmentation-based methods~\cite{mattyus2017deeproadmapper,batra2019improved,Hamza2022Semantic} can be integrated into the map generation pipeline for  efficiency improvement, these existing methods can only tackle a small fraction of element types, and fail to model elements with rather complex geometric shapes.
Differently, VMA models various map elements in a unified manner,
\ie, map elements with different geometric patterns (line-shaped, regular-shaped and closed-shaped) are all generally abstracted as point sequence for automatic annotation.
VMA  unifies and simplifies the  element representation. And this unified representation can be extended to most map  elements in the driving scene, and is compatible with various map standards (like OpenDRIVE~\cite{opendrive}, Lanelet2~\cite{Lanelet2}, and Apollo~\cite{apollo}).

Spatial extensibility is another key problem of map generation. 
On one hand, scaling up the spatial range of map is troublesome, due to  memory and computation limitations. On the other hand, some map elements (\eg, road curb) have a long spatial coverage. Keeping the continuity and completeness of these elements is difficult.
To solve the spatial extensibility problem, we design a divide-and-conquer annotation scheme and a vectorized merging algorithm, VMA has no restriction on the spatial range and is  widely applicable.

\begin{figure*}[]
    \centering
    \includegraphics[width=\linewidth]{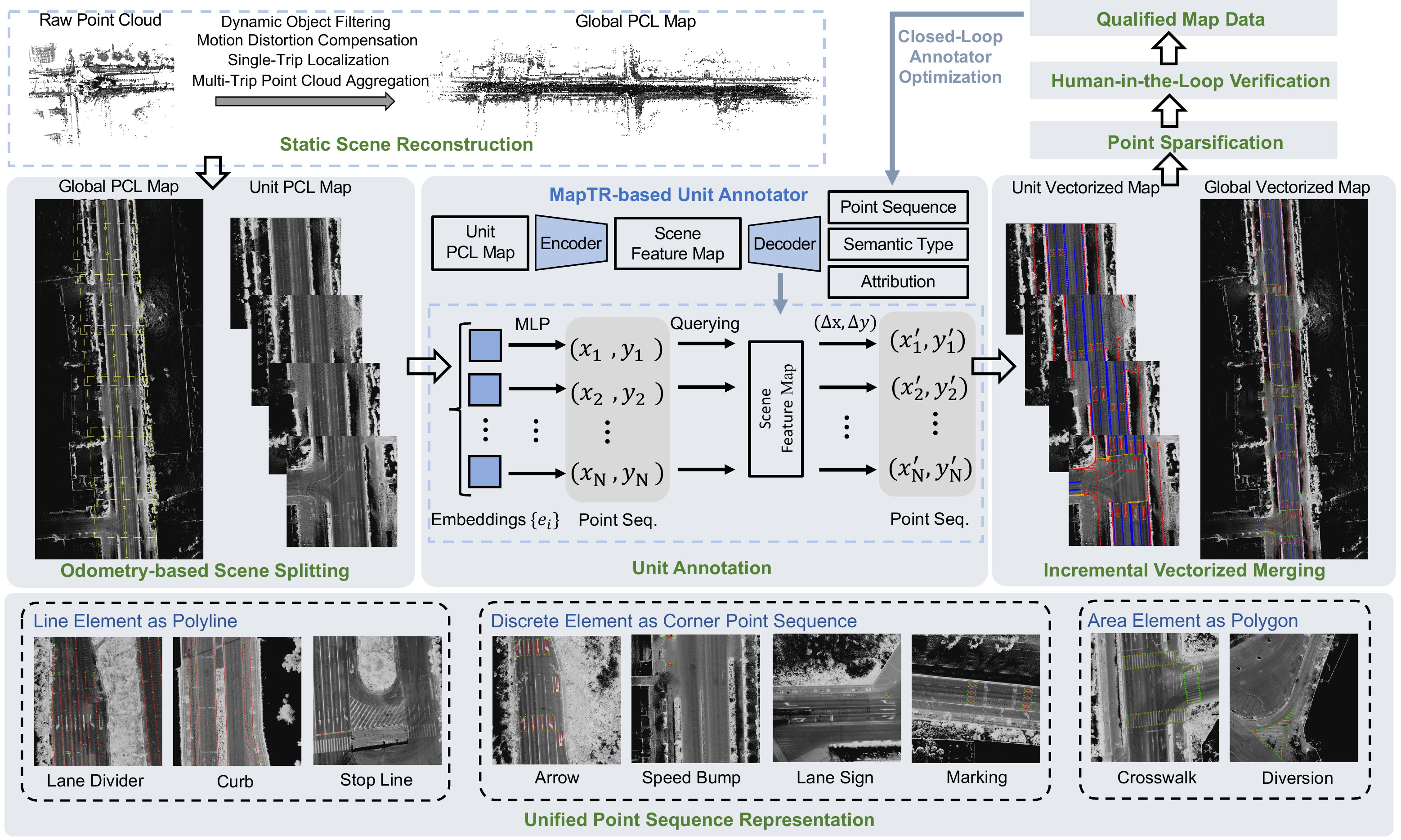}
    \caption{\textbf{The VMA framework}. VMA reconstructs the static scene through crowd-sourced multi-trip aggregation, and adopts a divide-and-conquer pipeline for scene annotation. VMA abstracts map elements with a variety of geometric patterns as unified point sequence representation for vectorized map annotation.}
    \label{fig:framework}
\end{figure*}

An overview of the proposed framework is shown in Fig.~\ref{fig:framework}. 
Specifically, VMA reconstructs the static scene through crowd-sourced multi-trip aggregation. Based on the reconstructed point cloud map (PCL map) and odometry information, VMA splits the scene into annotation units.
We build up MapTR-based~\cite{maptr} Unit Annotator model, which takes unit PCL map as input and directly outputs unit vectorized map.
Then vectorized map merging is performed to merge all unit vectorized maps into global vectorized map. Finally, point sparsification and human verification are performed to improve the annotation results.

VMA is highly automatic and significantly improves map generation efficiency.
To validate the effectiveness, we apply VMA in both urban scene and highway scene. On average, VMA takes 160min for annotating a scene with a range of hundreds of meters and reduces 52.3\% of the human cost.
We also leverage VMA to perform lane boundary detection on NYC Planimetric Database for benchmark evaluation.
VMA is of great application value in autonomous driving and can also be applied to other robotic scenarios.

This paper's contributions are summarized as follows:
\begin{itemize}
    \item  We design a divide-and-conquer annotation scheme to solve the spatial extensibility problem of HD map generation, and abstract map elements with a variety of geometric patterns as unified point sequence representation.
    \item Based on the unified map representation and divide-and-conquer annotation scheme, we build up a systematic vectorized map annotation framework (termed VMA) for automatically generating HD map of large-scale driving scenes. VMA is highly automatic and extensible, requiring negligible human effort, and flexible in terms of spatial scale and element type.
    \item We quantitatively and qualitatively validate the annotation performance on  real-world urban and highway scenes, as well as NYC Planimetric Database.
    VMA can significantly improve map generation efficiency and require little human effort, showing great industrial application value.
\end{itemize}

The remaining part of the paper is organized as follows.
A review of the related works on road marking extraction, line-shaped element extraction, online mapping, and lane detection is given in Section~\ref{sec:related_work}. Section~\ref{sec:framework} describes the overall design of VMA, including  unified point sequence representation, static scene reconstruction procedure, and divide-and-conquer scene annotation scheme.
Experiments are presented in Section~\ref{sec:experiment}. And finally, conclusions and discussions are presented in Section~\ref{sec:conclusion}.

\section{Related Work}  
\label{sec:related_work}

\subsection{Road Marking Extraction}
Road markings are signs on road surfaces usually painted with highly retro-reflective materials, which make them noticeable to human vision and sensors of autonomous vehicles. 
Road markings are essential features of HD map because they provide  rich traffic information (lane type,  lane direction,  \etc) for navigation.
Traditionally, road marking extraction is achieved through image processing methods~\cite{canny,Zhang2021Research}, \eg,
image denoising, image enhancement,
edge-based detection, k-mean clustering, and regional growth.
With the rapid development of deep learning, CNN-based methods  have been widely employed in detecting and recognizing road markings.
\cite{LaneNet} predicts  a semantic
segmentation binary mask to distinguish lane markings from the background.  
\cite{Franz2019Deep} designs a wavelet-enhanced FCNN to segment high-resolution aerial imagery, and create a 3D reconstruction of road markings based on the least-squares line-fitting.
\cite{MarkCapsNet} proposes a self-attention-guided capsule network, to extract road markings from aerial images. 
Different from these works, we model road marking as corner point sequence to achieve unified map element representation. 

\subsection{Line-shaped Element Extraction}
Existing methods for line-shaped element extraction can be divided into two kinds: segmentation-based methods~\cite{mattyus2017deeproadmapper,batra2019improved,Hamza2022Semantic} and iterative  methods~\cite{liang2019convolutional,xu2021icurb,enhancedicurb}.
Segmentation-based methods predict segmentation probabilistic map from aerial images and then perform heuristic post-processing algorithms on the segmentation map to get the line-shaped map representation.
DeepRoadMapper~\cite{mattyus2017deeproadmapper}  segments the aerial images into interest categories, and connects the endpoint of the discontinued roads based on  a specific distance threshold to alleviate the discontinuity issue of the road segmentation.
\cite{batra2019improved} fixes the road network disconnectivity issue by predicting the orientation and segmentation of the road network and correcting the segmentation results with n-stacked multi-branch CNN. \cite{Hamza2022Semantic} adopts the encoder-decoder architecture  as well as the attention mechanism, and introduces an edge detection algorithm to refine the segmented results.
Iterative graph growing methods predict the map element in an auto-regressive manner (vertex by vertex). \cite{liang2019convolutional} designs a convolutional recurrent network to predict the road boundary.
\cite{xu2021icurb,enhancedicurb} adopt better training strategy and graph growing policy to improve performance from the perspective of imitation learning.
VMA models both line-shaped elements and road markings, as well as area elements, in a unified vectorized manner.

\begin{figure*}[]
    \centering
    \includegraphics[width=\linewidth]{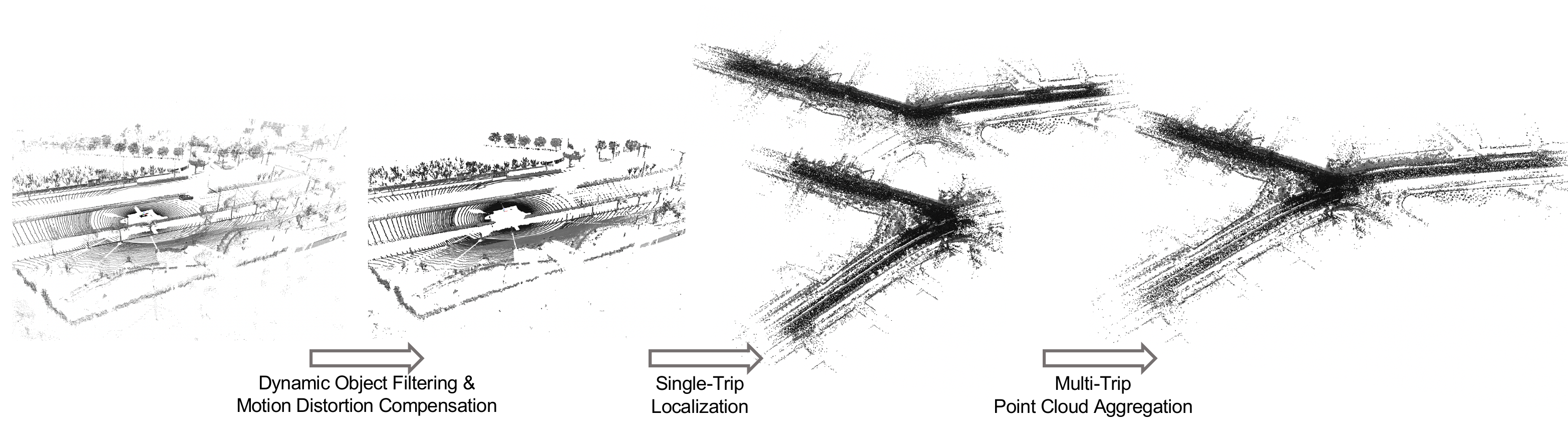}
    \caption{\textbf{Static scene reconstruction}. VMA generates dense point cloud map through crowd-sourced multi-trip aggregation. }
    \label{fig:scene_reconstruction}
\end{figure*}

\subsection{Online Mapping} Online mapping aims to create vehicle-centric local map on the fly and demands high efficiency.
With the development of PV-to-BEV transformation~\cite{Ma2022VisionCentricBP},
previous methods 
~\cite{polarbev,cvt,bevformer,liu2022bevfusion,liu2022petrv2,lu2022ego3rt,gkt} perform BEV semantic segmentation based on surround-view image data captured by vehicle-mounted cameras.
To build vectorized semantic map, HDMapNet~\cite{hdmapnet} adopts a segmentation-then-vectorization paradigm.
To achieve end-to-end learning~\cite{detr,deformdetr,yolos}, VectorMapNet~\cite{vectormapnet} 
adopts a coarse-to-fine two-stage pipeline and utilizes an auto-regressive decoder to predict points sequentially.
MapTR~\cite{maptr} proposes  permutation-equivalent modeling to exploit the undirected nature of semantic map and designs a parallel end-to-end framework. 
Online methods create vehicle-centric local map and focus on the balance of efficiency and accuracy.
VMA  aims at creating large-scale global map and focuses on map generation quality.

\subsection{Lane Detection}
Lane detection~\cite{tabelini2021keep,wang2022keypoint,garnett20193d,lstr,guo2020gen,liu2022learning, feng2022rethinking,chen2022persformer} is targeted at detecting lane elements in the scene.
LaneATT~\cite{tabelini2021keep} utilizes anchor-based deep lane detection model. LSTR~\cite{lstr} uses the Transformer architecture to output parameters of a lane shape model. GANet~\cite{wang2022keypoint} formulates lane detection as a keypoint estimation and association problem and takes a bottom-up design. \cite{garnett20193d} performs 3D lane detection in BEV. BezierLaneNet~\cite{feng2022rethinking} uses a fully convolutional network to predict 4-order Bezier lanes. PersFormer~\cite{chen2022persformer} 
proposes a Transformer-based architecture for spatial transformation and unifies 2D and 3D lane detection. 
VMA considers a wide range of map elements which include lane.

\section{Framework}
\label{sec:framework}
The framework of VMA is depicted in Fig.~\ref{fig:framework}.
VMA  first  performs static scene reconstruction (Sec.~\ref{sec:scene_reconstruction})  and
adopts a divide-and-conquer pipeline for scene annotation (Sec.~\ref{sec:scene_annotation}). The framework is highly automatic and requires little human effort.

\subsection{Static Scene Reconstruction}
\label{sec:scene_reconstruction}
The static scene reconstruction procedure is mainly composed of multi-trip data collection,  point cloud pre-processing (dynamic object filtering and motion distortion compensation), single-trip  localization and mapping, and multi-trip point cloud aggregation (Fig.~\ref{fig:scene_reconstruction}).

\noindent\textbf{Multi-Trip Data Collection.}
The data collection vehicles are equipped with multi-modal sensors (GPS, IMU, wheel, LiDAR and camera).
GPS, IMU, and wheel collect position and motion information. LiDAR generates point cloud map.
Camera collects color and texture information of the scene and is equipped for map verification.
The data collection procedure for a target scene is crowd-sourced and composed of several trips. On each trip, one vehicle takes a specific route travelling over the target scene.
Different vehicles take different routes to fully cover the target scene.   

\noindent\textbf{Dynamic Object Filtering.}
Since the scene reconstruction of VMA includes temporal aggregation, the motion of traffic participants (vehicles, pedestrians, cyclists, \etc{}) may result in noisy points in the point cloud map. Thus, before aggregation, VMA adopts 3D detection algorithm~\cite{afdetv2} to mark out objects with 3D bounding box and filter out points inside the box for each LiDAR frame. 

\noindent\textbf{Motion Distortion Compensation.}
LiDAR sensor scans the environment in a rotational manner. A frame (sweep) of point cloud is composed of a set  of scans at different timestamps of a rotation period.
When the ego vehicle is moving, the pose of LiDAR sensor is continuously changing within each period.
The scans of a frame are not aligned, resulting in motion distortion.
Based on IMU pre-integration, the ego motion is estimated and the motion distortion is calculated to de-skew LiDAR point clouds.

\noindent\textbf{Single-Trip Localization.}
In consideration of system efficiency,
firstly the localization~\cite{loam,legoloam,liosam,dynam-slam,orb-slam}  process is  independently and parallelly performed inside each trip.
We adopt LIO-SAM~\cite{liosam} for single-trip offline localization.  
Specifically,  initial odometry prior is generated through IMU pre-integration. Scan-matching~\cite{GICP} among LiDAR frames is used for odometry optimization. 
And GPS information is introduced to eliminate pose drift, which offers absolute pose measurements. 
Loop closure detection is performed to correct accumulated errors over time.

\noindent\textbf{Multi-Trip Point Cloud Aggregation.}
The localization results of different trips are further jointly optimized  through graph optimization~\cite{LT-mapper,poss_graph}.
The pose at each timestamp of each trip is defined as node of graph.
We adopt Scan-Context~\cite{scancontext} as frame descripter for ICP, and obtain the relative pose between nodes of different trips.
The relative pose of adjacent frames of the same trip is defined as the intra-trip edge, while the relative pose of two frames from two different trips is defined as the inter-trip edge.
With the inter-trip and intra-trip edge constraints, we complete the overall pose graph optimization and get the optimized localization results.
Based on the optimized localization results, all LiDAR frames are aligned for aggregation. The aggregated dense point cloud map  covers the whole target scene, and contains rich semantic information for scene annotation.

\subsection{Divide-and-Conquer Scene Annotation}  
\label{sec:scene_annotation}

\begin{figure}[]
    \centering
    \includegraphics[width=\linewidth]{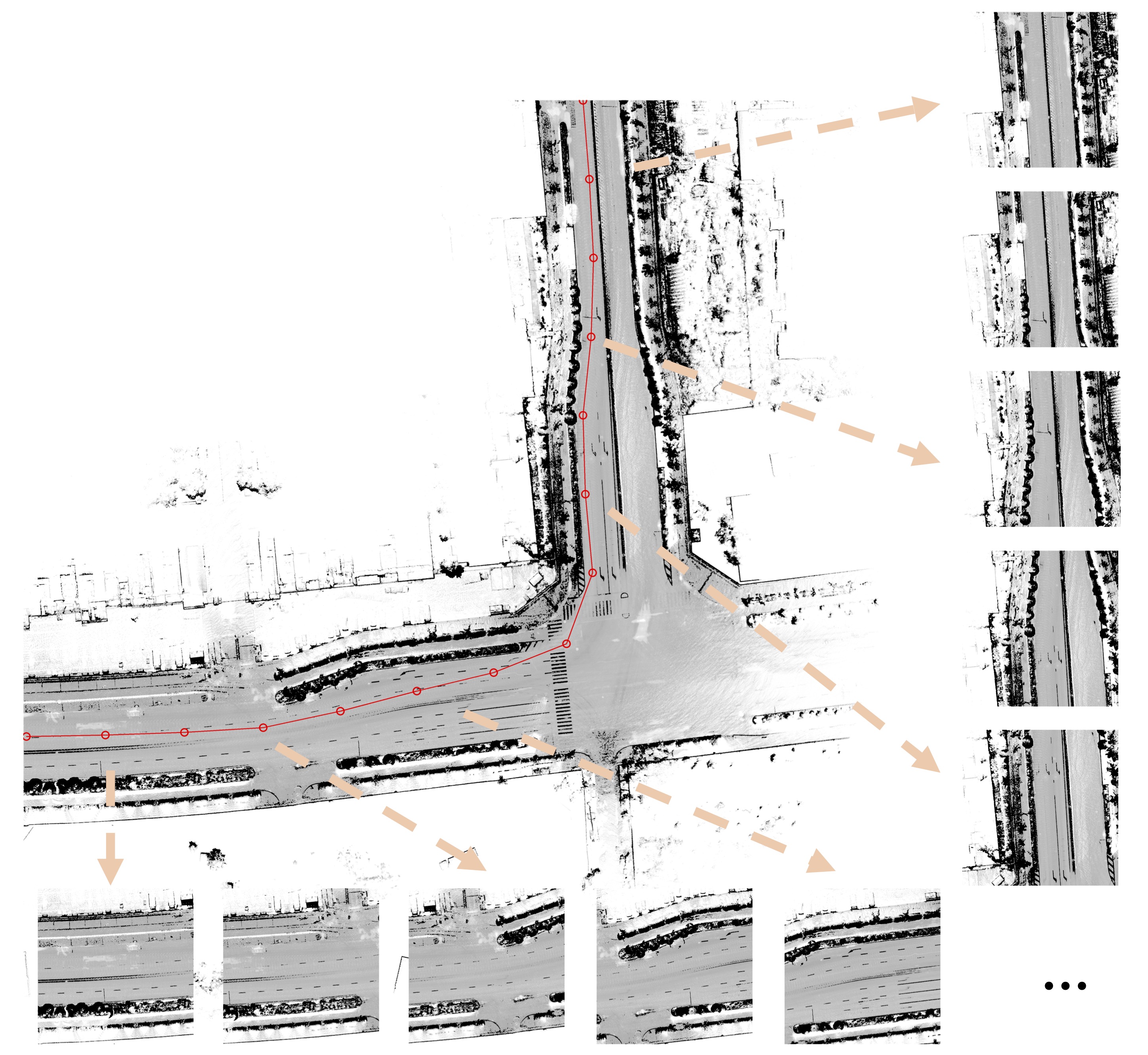}
    \caption{\textbf{Scene splitting}.  Ego trajectory and sampled ego position are respectively marked with red line and dot.}
    \label{fig:split}
\end{figure}
VMA is targeted at annotating spatially unlimited large-scale scene, for which one-shot annotation is infeasible due to memory and computation limitations.
Thus, a divide-and-conquer annotation scheme is designed to solve the spatial extensibility problem. Specifically,
VMA  splits the scene into overlapped annotation units, 
uses MapTR-based Unit Annotator to output the vectorized representation of map in each unit, and incrementally merges unit vectorized map to generate the global vectorized map.

\noindent\textbf{Unified Point Sequence Representation.}
VMA abstracts a wide range of map element as unified point sequence representation.
Based on the geometric characteristics,  map element is categorized into three main types, \ie, line element, discrete element and area element (see Table~\ref{tab:element-type} and Fig.~\ref{fig:framework}).

Line elements mainly include road curb, lane divider, and stop line. Through sampling points at a fixed interval along the line, we get the corresponding point sequence representation $\{p_i\}^N$ ($N$ denotes the point number). By sequentially connecting these points,  we get N-point polylines to represent line elements.

Discrete elements include all regular-shaped elements (like arrow, speed bump, and diamond marking), and is represented by the four corner points $\{p_i\}^4$ of the bounding box.  
The four points are ordered (front-left, front-right, back-right, and back left). The orientation of the element can be inferred from the order.

Area elements include closed-shaped regions, \eg, crosswalk, road intersection, diversion, \etc{}
We frame the annotation of these regions  as boundary regression, \ie, 
through sampling points at a fixed interval along the region's boundary,
we convert the boundary into an ordered point sequence $\{p_i\}^N$, which forms N-point polygons to represent the regions. 
We annotate the regions by detecting these points.

\begin{table*}
\caption{\textbf{Geometric type, vectorized representation, semantic type, and attribution of map element in VMA.} Map elements with different geometric patterns (line elements, discrete elements and area elements) are all abstracted as unified point sequence representation for vectorized map annotation. VMA also outputs attributions of map element.}
\centering
\resizebox{0.98\linewidth}{!}{
\begin{tabular}{l c c c}
\toprule
Geometric Type & Vectorized Representation & Semantic Type  & Attribution\\
\midrule
\multirow{4}{*}{Line Element} & \multirow{4}{*}{$N$-Point Sequence (Polyline)} &  Lane Divider  & Direction: Unidirectional / Bidirectional;  Line Type: Solid / Dotted / Fishbone; ...\\
& & Curb  & Curb Type: Ground Side / Road Side / Guardrail \\
& & Stop Line &  - \\
& & ... & ...\\
\midrule
\multirow{5}{*}{Discrete Element} & \multirow{5}{*}{Corner Point Sequence} & Arrow  & Arrow Type: Straight / Turn Off / Merge Right / No Turn Left / ...; ...\\
& & Speed Bump &  - \\
& & Lane Sign &  Lane Sign Type: Bike Lane / Bus Lane \\
& & Marking &   Marking Type: Diamond Marking / Inverted Triangle Marking\\
& & ... & ...\\
\midrule
\multirow{3}{*}{Area Element} & \multirow{3}{*}{$N$-Point Sequence (Polygon)} &  Crosswalk  & -\\
& & Diversion &  -\\
& & ... & ...\\

\bottomrule
\end{tabular}
}
\label{tab:element-type}
\end{table*}

\noindent\textbf{Odometry-based Scene Splitting.}
VMA splits the scene into fixed-sized annotation units based on the odometry information (Sec.~\ref{sec:scene_reconstruction}).
Specifically, as depicted in Fig.~\ref{fig:split}, along the vehicle's trajectory, VMA samples ego positions at fixed time or distance intervals, denoted as  $\mathbb{P} = \{\text{pos}_{1}, \text{pos}_{2}, ..., \text{pos}_{n}\}$.
Centered at these sampled positions, VMA crops unit PCL map from the global PCL map. And the scene is split into a group of annotation units along the ego trajectory.

\noindent\textbf{Unit Annotation.}
We build up MapTR-based Unit Annotator  model (see Fig.~\ref{fig:framework}) which takes unit PCL map as input and directly outputs unit vectorized map.
The unit PCL map is first sent into an encoder network, which flattens the unit PCL map into 2D map, extracts features with CNN, and outputs 2D scene feature map.
Then a DETR-like~\cite{detr,maptr} decoder predicts map elements in a set-to-set manner. Each element corresponds to one sequence of learnable embeddings $\{e_i\}^N$ (one embedding for one point). Each embedding is sent into a MLP layer to predict point's coordinate $(x, y)$ in the scene feature map. And then we sample features at $(x, y)$ of 2D scene feature map for predicting offsets $(\Delta x, \Delta y)$  to update $(x, y)$.  
Feature sampling and coordinate update are iteratively performed. Finally, the decoder outputs the point sequence represention of map element, as well as corresponding semantic type and attribution (see Table~\ref{tab:element-type}).

\begin{figure}[]
    \centering
    \includegraphics[width=\linewidth]{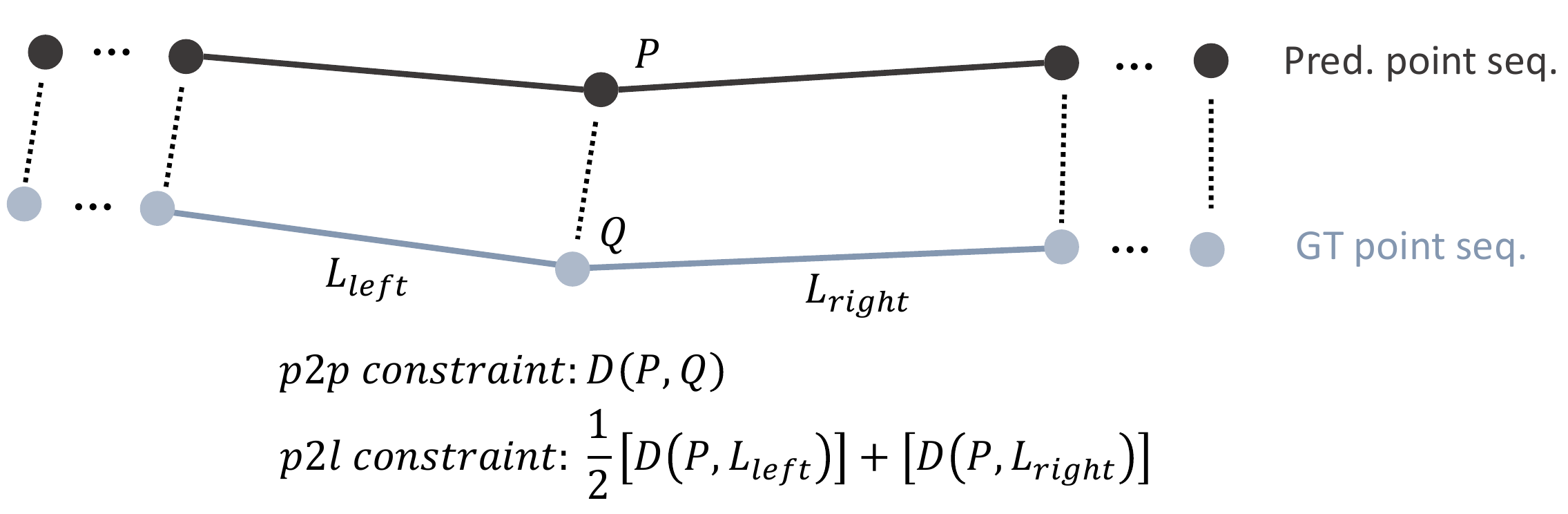}
    \caption{\textbf{Geometric constraint of map element.}}
    \label{fig:p2l-p2p}
\end{figure}

The MapTR-based Unit Annotator is initially optimized with a small set of human-annotated scene data,
and continuously optimized in a closed-loop manner (see Sec.~\ref{sec:human-verification}).
The human-annotated labels are converted to the vectorized point sequence representation for supervision.
In the training phase, we perform one-to-one hierarchical assignment~\cite{maptr} between human-annotated elements and predicted elements.
After one-to-one hierarchical assignment, the predicted point sequence is paired with the GT point sequence point by point (as depicted in Fig.~\ref{fig:p2l-p2p}, $P$ matches $Q$).  
Two kinds of supervision are adopted to constraint the geometry, p2l (point-to-line)  and p2p (point-to-point) constraint.
P2p constraint is the $L_2$ distance between paired points,
while the P2l constraint is the point-to-line distance between predicted point $P$ and the two adjacent edges of $Q$.
P2p constraint is applied to key points which require exact localization, \ie, the start and end points of line element, and corner points of discrete elements, while p2l constraint is applied to other points for adaptively keeping the geometry. 
When $L_{left}$ and $L_{right}$ are non-collinear, p2l constraint is equivalent to p2p constraint and makes $P$ converge to $Q$; when $L_{left}$ and $L_{right}$ are collinear, under the p2l constraint $P$ is not necessary to converge to $Q$ but is on the line.  In this case, the convergence constraint is relaxed and also keeps the same geometry of element.
Semantic type and attribution of each element are predicted by a MLP layer and supervised with cross-entropy loss.

\begin{figure}[]
    \centering
    \includegraphics[width=\linewidth]{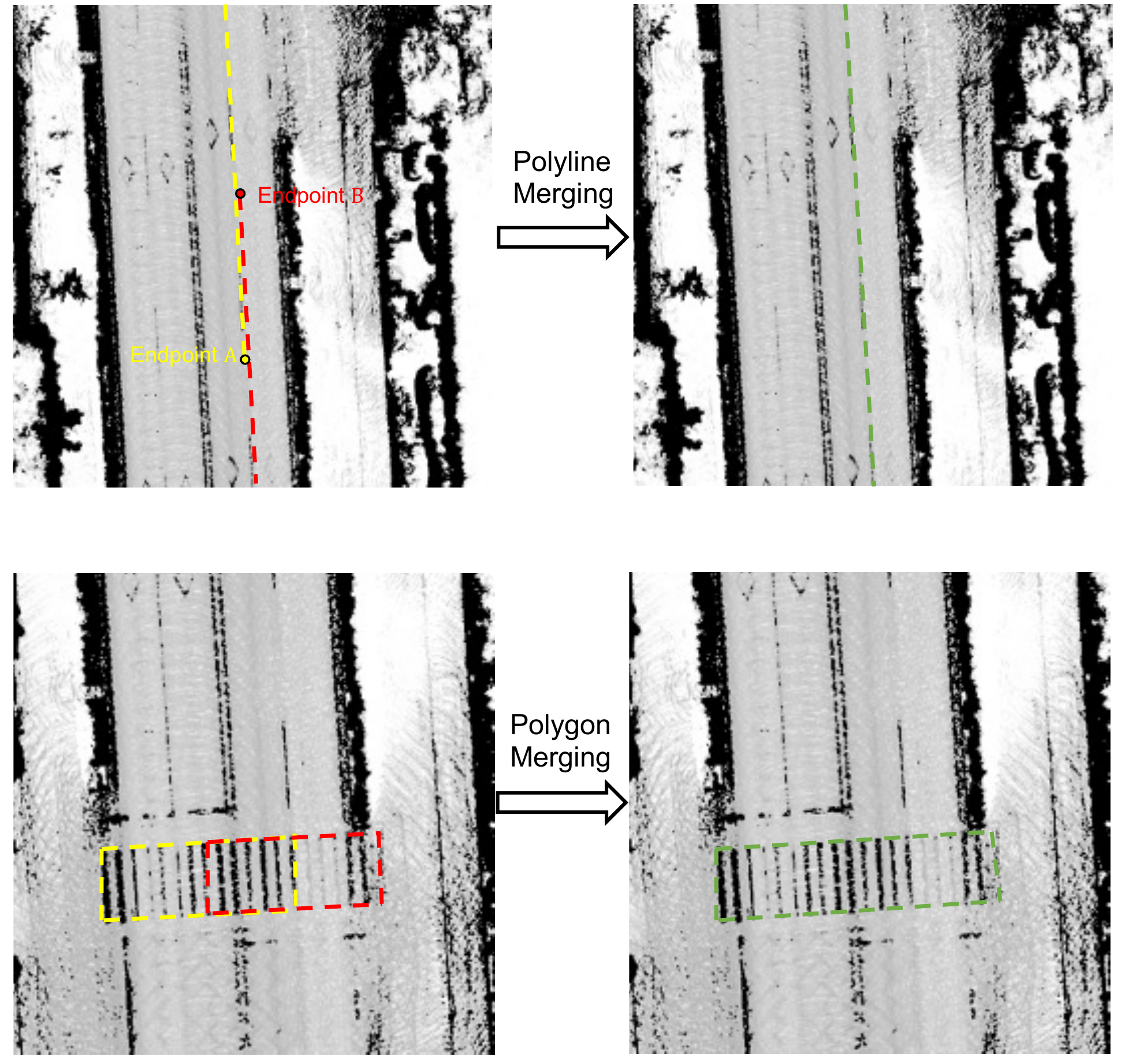}
    \caption{\textbf{Illustration of merging strategies for  line element (polyline) and area element (polygon).}}
    \label{fig:vectorized-merging}
\end{figure}

\noindent\textbf{Incremental Vectorized Map Merging.}
MapTR-based Unit Annotator outputs unit vectorized map, consisting of fragmented map elements because of the unit border truncation.
Incremental vectorized merging is performed to merge all unit vectorized maps $\{M^i_{local}\}$ into global vectorized map $M_{global}$.
Concretely, based on the ego trajectory, unit vectorized maps are incrementally and sequentially merged:
\begin{equation}
    M_{global} = (...(((M^1_{local} \oplus M^2_{local}) \oplus M^3_{local}) \oplus M^4_{local})  ...\oplus M^n_{local}),
\end{equation}
where $\oplus$ denotes the merging process.
The merging strategies vary according to the geometric patterns of map element:
\begin{itemize}
    \item Line element (polyline): line elements are associated if the following conditions are met (refer to Fig.~\ref{fig:vectorized-merging} (top)): two polylines overlap with each other; the overlapped length exceeds a threshold $\theta_{line}$; the endpoints of two polylines are close to the other polyline.
   We merge the associated elements (red one and yellow one)  into one polyline (green one) by simply removing the overlapped part.
    \item Discrete element (corner point sequence). Chamfer distance $D_{chamfer}$ between two corner point sequences are calculated.  If the distance is smaller than a threshold $\delta_{discrete}$, the two elements are associated. Through non-maximum suppression, associated elements are merged into one.
    \item Area element (polygon).  If the IoU between two polygons exceeds a threshold $\delta_{area}$, the two polygons are associated. Associated polygons are merged into one polygon with the union operation (see Fig.~\ref{fig:vectorized-merging} (bottom)).
    \item Attribution. 
    The merging strategy of attribution is majority vote.
     The attribution of merged element is the majority of all associated elements.
\end{itemize}

\begin{figure*}[]
    \centering
    \includegraphics[width=\linewidth]{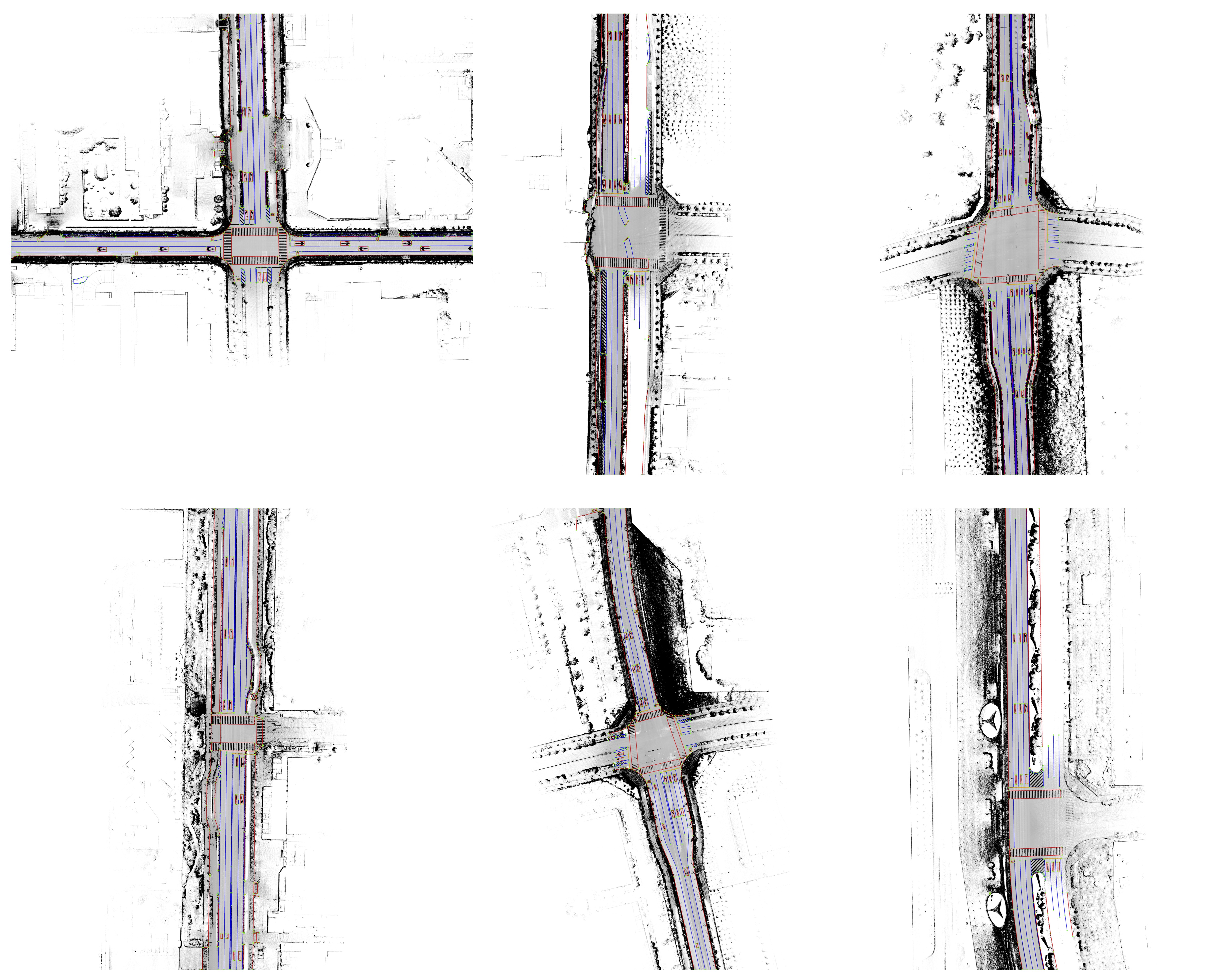}
    \caption{\textbf{Automatic annotation results of urban scene.} Zoon in for better view.}
    \label{fig:urban}
\end{figure*}

\noindent\textbf{Point Sparsification.}
After incremental vectorized merging, fragmented map elements are transformed into complete map elements which are represented by dense and redundant point sequence. For the convenience of application and storage efficiency, VMA adopts the Douglas-Peucker algorithm to simplify the representation of map element, which iteratively removes points from the polyline (or polygon) while keeping the geomety.

\noindent\textbf{Human-in-the-Loop Verification.}
\label{sec:human-verification}
Through the proposed divide-and-conquer scene annotation procedure, global vectorized map is obtained in a highly automatic manner. VMA introduces human verification for guaranteeing the annotation quality. After human verification, qualified vectorized map data are archived and regarded as extra training data for finetuning the MapTR-based Unit Annotator. With VMA running for a longer time, the annotation quality is continuously improved and less human effort is needed.

\section{Experiments}
\label{sec:experiment}

\subsection{Datasets}
\noindent\textbf{Real-World Urban Scene and Highway Scene.}
To  validate the effectiveness of VMA, we apply VMA to  map automatic annotation in both urban scene and highway scene.
We  collect a large amount of real-world scene data in a crowd-sourced way with the data collection vehicles equipped with multi-modal sensors. 
And the reported metrics are based on a validation set consisting of 850 urban scenes and 2082 highway scenes.
10231 urban scenes and 8334 highway scenes with  map annotations are used  to optimize the MapTR-based Unit Annotator.

\noindent\textbf{NYC Planimetric Database.}
For fair comparisions with methods based on aerial images, following the Topo-boundary benchmark~\cite{enhancedicurb}, we perform lane boundary detection on NYC Planimetric Database.
NYC Planimetric Database contains 2147 high-resolution aerial images ($5000\times5000$) with road-boundary annotations. 
Each pixel corresponds to $15.2cm$ in 3D space.
According to the Topo-boundary benchmark~\cite{enhancedicurb},
the high-resolution images are spilt into $1000\times1000$ images for evaluation. More details about the benchmark setting are available in \cite{enhancedicurb}.

\subsection{Experimental Settings}
For line and area element, every instance corresponds to a 50-point sequence (\ie, $N=50$).
We train the MapTR-based Unit Annotator of VMA on eight RTX 3090 GPUs with batch size 8.  By default, the spatial range of annotation unit is $50m \times 50m$ for urban and highway scenes and $1000$ pixels $\times$ $1000$ pixels for NYC Planimetric Database.
We use AdamW~\cite{loshchilov2017adamw} optimizer and Cosine Annealing~\cite{loshchilov2016cosineanneal} scheduler with weight decay 0.01 and initial learning rate $2\times{10}^{-4}$.

\begin{figure*}[]
    \centering
    \includegraphics[width=\linewidth]{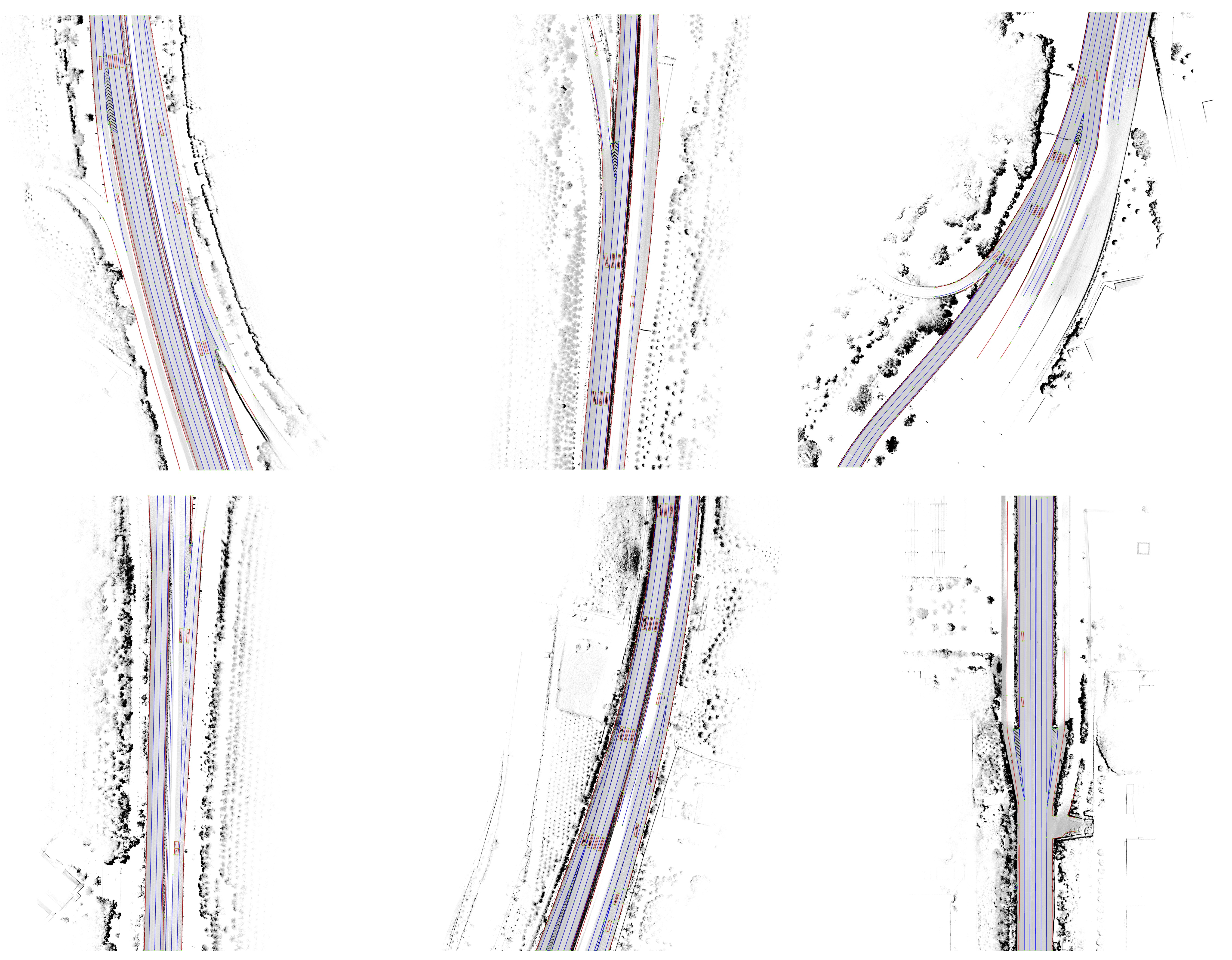}
    \caption{\textbf{Automatic annotation results of highway scene.} Zoon in for better view.}
    \label{fig:highway}
\end{figure*}

\subsection{System Runtime}
VMA is a highly automatic map annotation system. We present the detailed runtime of the system in Table~\ref{tab:runtime}.  
For static scene reconstruction,   
the data collection is performed crowd-sourced way and approximately takes 30min per scene.
Dynamic object filtering, single-trip localization, and multi-trip point cloud aggregation respectively takes 15min, 40min, and 60min per scene on average.
Thanks to the divide-and-conquer scheme, the scene annotation procedure is quite efficient. Unit annotation is parallelly performed on GPU and costs little time.
VMA requires negligible human effort (12min/scene averagely for map quality verification). And the overall runtime for annotating a scene (with a range of hundreds of  meters) is 160min on average.

\begin{table*}
\caption{\textbf{Detailed runtime of VMA.}}
\centering
\resizebox{0.8\linewidth}{!}{
\begin{tabular}{l c c c c c}
\toprule
Procedure & Sub-procedure & Time & Hardware\\
\midrule
\multirow{5}{*}{Scene Reconstruction} &  Data Collection & $\approx$ 30min/scene & -\\
 &  Dynamic Object Filtering & $\approx$ 15min/scene & GPU \& CPU\\
 & Motion Distortion Compensation & $\leq$ 1min/scene & CPU\\
  &  Single-Trip Localization & $\approx$ 40min/scene & CPU\\
  &  Multi-Trip Point Cloud Aggregation & $\approx$ 
60min/scene & CPU \\ 
\midrule
\multirow{5}{*}{Scene Annotation} & Scene Splitting &  $\leq$ 1min/scene & CPU \\
& Unit Annotation & $\leq$ 1min/scene & GPU  \\
& Vectorized Merging & $\leq$ 1min/scene & CPU \\
& Point Sparsification & $\leq$ 1min/scene & CPU \\
& Human Verification & $\approx$ 12min/scene & - \\
\midrule
Overall & - &  $\approx$160min/scene  & -\\
\bottomrule
\end{tabular}
}
\label{tab:runtime}
\end{table*}

\begin{table*}[] 
\caption{\textbf{Quantitative results on NYC Planimetric Database for comparison.} The best results are highlighted in bold font. For all the metrics, larger values indicate better performance.} 
\centering 
\resizebox{0.98\linewidth}{!}{
\begin{tabular}{l c c c c c c c c c c c c c c c c}
\toprule
\multirow{2}{*}{Methods}& \multicolumn{3}{c}{Precision $\uparrow$} & \multicolumn{3}{c}{Recall $\uparrow$} & \multicolumn{3}{c}{F1-score $\uparrow$} & \multirow{2}{*}{Naive $\uparrow$} & \multirow{2}{*}{APLS $\uparrow$} & \multirow{2}{*}{ECM $\uparrow$} \\ 
\cmidrule(l){2-4} \cmidrule(l){5-7} \cmidrule(l){8-10} &  2 pixels &  5 pixels &  10 pixels &  2 pixels &  5 pixels &  10 pixels &  2 pixels &  5 pixels &  10 pixels \\
\midrule
Seg.-then-skeleton.~\cite{xu2021icurb} & 0.607&\textbf{0.890}&0.928&0.505&0.736&0.768&0.533&0.778&0.811& 0.698& 0.577& 0.550 \\
Deeproadmapper~\cite{mattyus2017deeproadmapper} & 0.578&0.854&0.898&0.475&0.694&0.725&0.505&0.740&0.775 & 0.719 & 0.615 & 0.595\\ 
OrientationRefine~\cite{batra2019improved} &\textbf{0.620}&0.878&0.913&\textbf{0.602}& 0.850 & 0.884 &\textbf{0.605}& 0.855& 0.888 & 0.797& 0.750& 0.756\\ 
RoadTracer~\cite{bastani2018roadtracer} & 0.391&0.707&0.791&0.416&0.743&0.821&0.399&0.718&0.798& 0.869& 0.739& 0.824\\ 
VecRoad~\cite{tan2020vecroad}&0.461&0.769&0.854&0.459&0.752&0.830&0.458&0.756&0.837& 0.883& 0.756& 0.846\\ 
ConvBoundary~\cite{liang2019convolutional} &  0.510 &0.845& 0.934 & 0.455&0.692&0.752& 0.465&0.737&0.805&\textbf{0.958}& 0.671& 0.786\\ 
DAGMapper~\cite{homayounfar2019dagmapper} & 0.407 & 0.751 & 0.868 & 0.353 & 0.649 & 0.747 & 0.371 & 0.684 & 0.787 & 0.896 & 0.679 & 0.758\\ 
iCurb~\cite{xu2021icurb} & 0.550&0.833&0.890&0.538&0.815&0.873&0.542&0.820&0.877&0.910 & 0.826 & 0.889 \\
Enhanced-iCurb~\cite{enhancedicurb} & 0.560&0.839&0.894&0.542&0.811&0.864&0.549&0.821&0.874&0.925& 0.822& 0.893\\
VMA & 0.533 & 0.872 & \textbf{0.935} & 0.530 & \textbf{0.856} & \textbf{0.913} & 0.528 & \textbf{0.858} & \textbf{0.916} & 0.900 & \textbf{0.865} & \textbf{0.901} \\
\bottomrule 
\label{tab:NYC}
\end{tabular}}
\end{table*}

\subsection{Human Cost Comparison}
As shown in Table~\ref{tab:humancost}, on average, manual annotation requires 25min per scene. VMA requires 12min per scene and reduces 52.3\% of the human cost. 

\begin{table}
\caption{\textbf{Human Cost Comparison.}}
\centering
\resizebox{0.9\linewidth}{!}{
\begin{tabular}{l c }
\toprule
Annotation Manner & Human Cost \\
\midrule
Manual Annotation & $\approx$25min/scene \\
Auto Annotation w/ Verification (VMA) &  $\approx$12min/scene \\
\bottomrule
\label{tab:humancost}
\end{tabular}
}
\end{table}

\subsection{Qualitative Evaluation}
\label{sec:qualitative}
\begin{figure*}[]
    \centering
    \includegraphics[width=\linewidth]{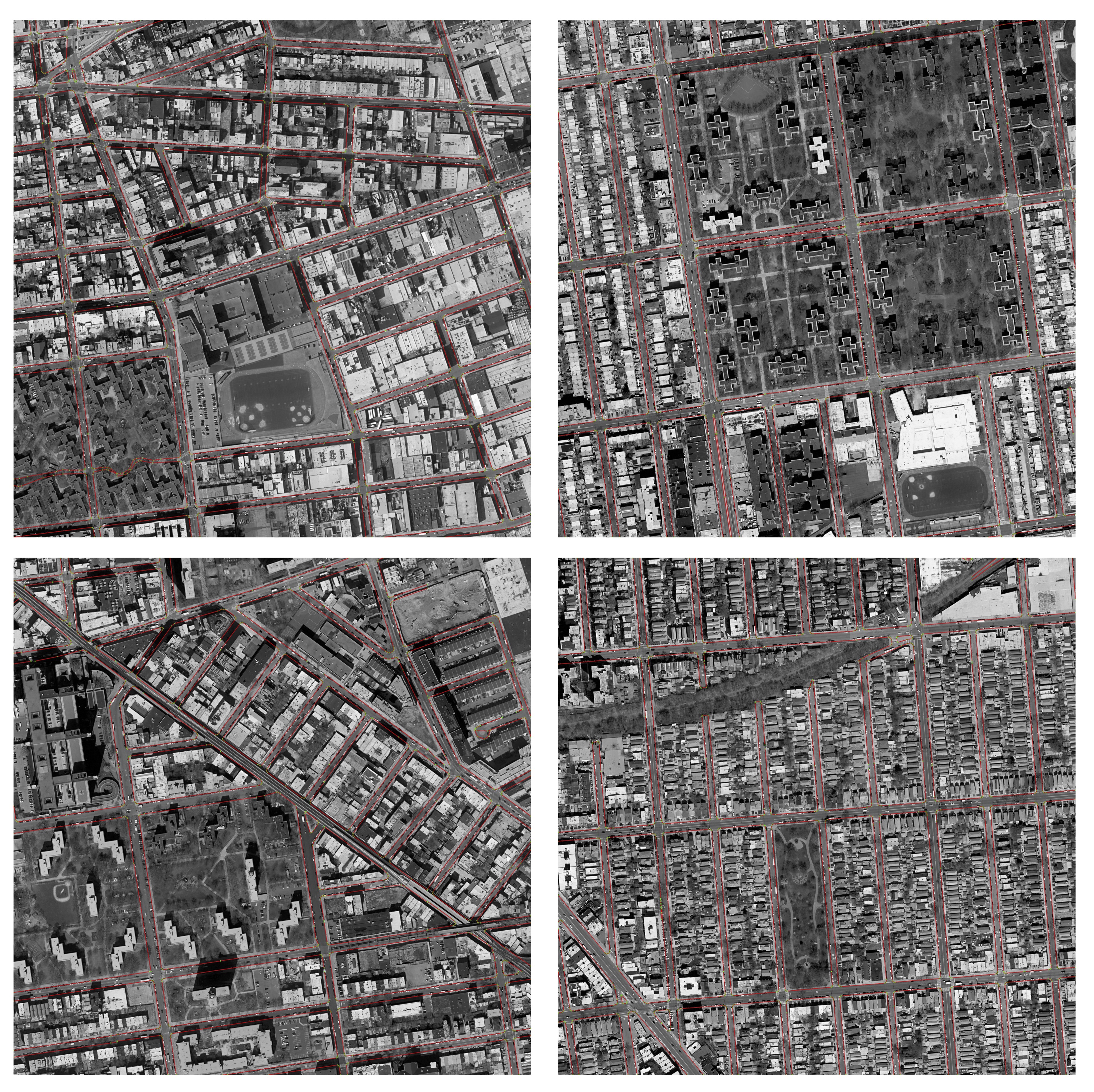}
    \caption{\textbf{Automatic annotation results of NYC Planimetric Database.} Zoon in for better view.}
    \label{fig:NYC}
\end{figure*}

Qualitative results of urban scene, highway scene, and NYC Planimetric Database are presented in Fig.~\ref{fig:urban}, Fig.~\ref{fig:highway}, and Fig.~\ref{fig:NYC}. 
VMA directly outputs high-quality global vectorized map for the whole scene, through a highly automatic divide-and-conquer scene annotation procedure.
Human annotators only need to verify  the automatic annotation results and  adjust a small fraction of map elements.
The map generation efficiency is significantly improved.   

We can also observe some failed cases in Fig.~\ref{fig:urban} and Fig.~\ref{fig:highway}. 
Specifically, if the  point cloud map is not dense enough and the scene features are not rich enough, the map annotation quality is not satisfactory.  The solution to this problem is increasing the trip number of data collection for better covering the target scene.

\subsection{Quantitative Evaluation}
\label{sec:quantitative}

We use the  same metrics used in Topo-boundary benchmark~\cite{enhancedicurb}  for quantitative comparison, which include pixel-level metrics (precision, recall, and F1-score), Naive Connectivity~\cite{liang2019convolutional}, APLS (Average
Path Length Similarity)~\cite{SpaceNet}, and ECM (Entropy-based Connectivity)~\cite{enhancedicurb}.

\noindent\textbf{Pixel-level Metrics.} Pixel-level metrics compose of precision, recall and F1-score. For pixel-level evaluation,  predicted map elements and ground truth elements are firstly densified into rasterized representation. Suppose $P_{\text{pre}}$ is the densified pixel set of predicted map elements and $P_{\text{gt}}$ is the densified pixel set of ground truth map elements. Then, the Euclidean distance of every pixel pair between $P_{\text{pre}}$ and $P_{\text{gt}}$ are calculated. Denote $cd_{\text{pre}}$ as the chamfer distance of one predicted pixel to $P_{\text{gt}}$ and $cd_{\text{gt}}$ as the chamfer distance of one ground truth pixel to $P_{\text{pre}}$. Precision is the ratio of predicted pixels whose $cd_{\text{pre}}$ are smaller than preset threshold $\tau$ in all predicted pixels. Recall is the ratio of ground truth pixels whose $cd_{\text{gt}}$ are smaller than preset threshold $\tau$ in all ground truth pixels. The metrics are formulated as:
\begin{equation} 
\begin{aligned}
        &\text{Precision} = \frac{|\{
        p | d(p,P_{\text{gt}})<\tau,\forall p\in P_{\text{pre}}
        \}|}{|P_{\text{pre}}|},\\
        &\text{Recall} = \frac{|\{
        p | d(p,S_{\text{pre}})<\tau,\forall p\in P_{\text{gt}}
        \}|}{|P_{\text{gt}}|},\\
        &\text{F1-score} =\frac{2 \cdot \text{Pecision} \cdot \text{Recall}}{\text{Precision} + \text{Recall}}.
\end{aligned}
\end{equation}

\noindent\textbf{Naive Connectivity.} This metric measures the connectivity of predicted instance. Naive Connectivity uses the Hausdorff distance to match every predicted instance and ground-truth instance. Every predicted instance will be assigned to one ground-truth instance, according to the smallest Hausdorff distance. Multiple predicted instances could be assigned to one ground-truth instance. $M_{i}$ represents the number of predicted instances to which each true value is assigned. $C_i=\frac{1(M_i>0)}{M_i}$ is the connectivity of ground-truth, which means the short prediction of long ground-truth instance will be punished. The final result is the average sum of connectivity of each ground-truth instance.

\noindent\textbf{APLS.} 
APLS (Average
Path Length Similarity) is proposed by \cite{SpaceNet} and has been widely used to evaluate topology correctness. It's based on Dijkstra's algorithm to measure the similarity of path.

\begin{table}[t] 
\setlength{\abovecaptionskip}{0pt} 
\setlength{\belowcaptionskip}{0pt} 
\centering 
\caption{\textbf{Comparisons of training and inference time on NYC Planimetric Database.}} 
\resizebox{0.8\linewidth}{!}{
\begin{tabular}{l c c}
\toprule
Method & Training & Inference \\
\midrule
Seg.-then-skeleton.~\cite{xu2021icurb} & 0.45s/scene&0.33s/scene \\ 
Deeproadmapper~\cite{mattyus2017deeproadmapper}&1.32s/scene&1.57s/scene\\ 
OrientationRefine~\cite{batra2019improved}&1.14s/scene&0.96s/scene \\ 
RoadTracer~\cite{bastani2018roadtracer}&6.50s/scene &4.92s/scene \\ 
VecRoad~\cite{tan2020vecroad}&19.89s/scene&14.17s/scene\\ 
ConvBoundary~\cite{liang2019convolutional} & 17.47s/scene &  25.17s/scene\\ 
DAGMapper~\cite{homayounfar2019dagmapper} & 8.19s/scene &  4.92s/scene\\ 
iCurb~\cite{xu2021icurb} & 6.74s/scene &  3.11s/scene\\
Enhanced-iCurb~\cite{enhancedicurb} & 7.25s/scene & 2.38s/scene\\
VMA & \textbf{0.35s/scene} & \textbf{0.11s/scene}\\
\bottomrule 
\label{tab:time}
\end{tabular}}
\end{table}

\noindent\textbf{ECM.}  Naive Connectivity ignores the length of predicted instances.  ECM (Entropy-based Connectivity) metric is used in \cite{enhancedicurb}, which assigns predicted longer instances with higher weight in the metric.
ECM is formulated as:
\begin{equation}
\begin{aligned}
&ECM = \sum_{i=1}^N \alpha_i e^{-C_i}, \\
&C_i = \sum_{j=1}^{M_i}-p_jlog(p_j), \\
&p_j = \frac {\text{length}(G^j_{\text{pre}})}{\sum_{I\in S_{i}} \text{length}(I)}. \\
\end{aligned}
\end{equation}
In ECM, Hausdorff distance matching is replaced with pixel-level voting mechanism. Each pixel of predicted instance votes ground-truth instance with the closest Euclidean distance, and the predicted instance will be assigned to ground-truth instance with the most votes. And the connectivity value $C_i$ is calculated from the entropy of dominance value.  $S_{i}$ is the set of all predicted instances assigned to one ground truth $G^i_{\text{gt}}$. Dominance value $p_j$ is the ratio of the length of predicted instance $G^j_{\text{pre}}$ to the sum of the length of all the instance in $S_{i}$. $M_i$ is the number of $S_{i}$. 
$\alpha_i$ is the completion of $G_{gt}^i$, which is equal to the sum of the length of assigned instances in $G_{pre}$ projected onto $G_{gt}^i$ divided by the length of $G_{gt}^i$.

Quantitative  comparisons on  NYC Planimetric Database are presented in Table~\ref{tab:NYC}.  
VMA achieves the best results in terms of most metrics. Besides, VMA is efficient in both  training and inference phase (on average 0.35s/scene for training and 0.11s/scene for inference), as shown in Table~\ref{tab:NYC}.

Quantitative evaluations of urban scene are presented in Table~\ref{tab:urban}.
For all types of map elements, even under the strict distance threshold $0.30m$, VMA  achieves high  recall, precision, and F1-score. 
It shows the unified point sequence representation well models map elements with various geometric patterns.  
Especially for lane divider and curb, which are the most important map elements of the driving scene,  recall$@0.30m$, precision$@0.30m$, and F1-score$@0.30m$  all exceed $0.90$.
Some infrequent elements (like speed bump) correspond to limited training samples. With the training samples accumulated and the closed-loop annotator optimization, the annotation quality can be continuously improved.

Quantitative evaluations of  highway scene are presented in Table~\ref{tab:highway}. The metrics of the highway scene are much higher than those of the urban scene.  The highway scene is standardized and easy to annotate, while the urban scene is  more complex and includes a large number of intersection scenarios. This difference accounts for the gap of metrics.

Annotation accuracy of attributions is shown in Table~\ref{tab:attribution}. Attribution predictions of VMA are  quite accurate, requiring little human effort for correction.

\begin{table*}[] 
\setlength{\abovecaptionskip}{0mm} 
\setlength{\belowcaptionskip}{0mm} 
\caption{\textbf{Quantitative results of urban scene.}} 
\centering 
\resizebox{0.8\linewidth}{!}{
\begin{tabular}{l c c c c c c c c c c c c c c c c }
\toprule
\multirow{2}{*}{Semantic Type}& \multicolumn{3}{c}{Precision $\uparrow$} & \multicolumn{3}{c}{Recall $\uparrow$} &  \multicolumn{3}{c}{F1-score $\uparrow$}\\ 
\cmidrule(l){2-4} \cmidrule(l){5-7} \cmidrule(l){8-10} 
&  0.30m &  0.75m & 1.50m & 0.30m &  0.75m & 1.50m
& 0.30m &  0.75m & 1.50m & \\
\midrule
Lane Divider & 0.913 & 0.926 & 0.932 & 0.930 & 0.937 & 0.940 & 0.915 & 0.925 & 0.930 \\
Curb & 0.910 & 0.929 & 0.939 & 0.917 & 0.934 & 0.941 & 0.906 & 0.924 & 0.933\\
Stop Line & 0.780 & 0.886 & 0.922 & 0.780 & 0.887 & 0.918 & 0.767 & 0.874 & 0.907\\
Arrow & 0.764 & 0.857 & 0.870 & 0.810 & 0.908 & 0.919 & 0.769 & 0.862 & 0.873\\
Speed Bump & 0.632 & 0.828 & 0.883 & 0.626 & 0.806 & 0.860 & 0.607 & 0.795 & 0.850\\
Lane Sign & 0.682 & 0.934 & 0.951 & 0.651 & 0.904 & 0.926 & 0.652 & 0.900 & 0.919\\
Marking & 0.799 & 0.913 & 0.926 & 0.777 & 0.883 & 0.899 & 0.773 & 0.880 & 0.899\\
Crosswalk & 0.559 & 0.720 & 0.810 & 0.616 & 0.797 & 0.893 & 0.570 & 0.736 & 0.827\\
Diversion & 0.581 & 0.722 & 0.811 & 0.647 & 0.808 & 0.904 & 0.596 & 0.742 & 0.832\\
\midrule
Average & 0.736 & 0.879 & 0.894 & 0.750 & 0.873 & 0.911 & 0.728 & 0.847 & 0.886\\

\bottomrule 
\label{tab:urban}
\end{tabular}}
\end{table*}

\begin{table*}[] 
\setlength{\abovecaptionskip}{0mm} 
\setlength{\belowcaptionskip}{0mm} 
\caption{\textbf{Quantitative results of highway scene.}} 
\centering 
\resizebox{0.8\linewidth}{!}{
\begin{tabular}{l c c c c c c c c c c c c c c c c c}
\toprule
\multirow{2}{*}{Semantic Type}& \multicolumn{3}{c}{Precision $\uparrow$} & \multicolumn{3}{c}{Recall $\uparrow$} &  \multicolumn{3}{c}{F1-score $\uparrow$}  \\ 
\cmidrule(l){2-4} \cmidrule(l){5-7} \cmidrule(l){8-10} 
&  0.30m &  0.75m & 1.50m & 0.30m &  0.75m & 1.50m
& 0.30m &  0.75m & 1.50m & \\
\midrule
Lane Divider & 0.934 & 0.940 & 0.944 & 0.948 & 0.953 & 0.955 & 0.935 & 0.941 & 0.944 \\
Curb  & 0.882 & 0.912 & 0.925 & 0.910 & 0.937 & 0.948 & 0.887 & 0.916 & 0.928\\
Stop Line & 0.802 & 0.903 & 0.932 & 0.793 & 0.897 & 0.925 & 0.788 & 0.890 & 0.918\\
Arrow & 0.860 & 0.955 & 0.967 & 0.855 & 0.949 & 0.961 & 0.852 & 0.945 & 0.957 \\
Speed Bump & 0.694 & 0.866 & 0.906 & 0.648 & 0.841 & 0.889 & 0.654 & 0.837 & 0.882\\
Lane Sign  & 0.849 & 0.962 & 0.969 & 0.828 & 0.936 & 0.946 & 0.829 & 0.938 & 0.947\\
Marking & 0.919 & 0.981 & 0.988& 0.913 & 0.973 & 0.982& 0.913 & 0.973 & 0.982\\
Crosswalk & 0.685 & 0.831 & 0.927 & 0.677 & 0.821 & 0.913 & 0.673 & 0.816 & 0.910 \\
Diversion & 0.700 & 0.834 & 0.924 & 0.699 & 0.827 & 0.913 &0.690 & 0.820 & 0.908 \\
\midrule
Average & 0.814 & 0.909 & 0.932 & 0.808 & 0.904 & 0.936 & 0.802 & 0.897 & 0.931\\
\bottomrule 
\label{tab:highway}
\end{tabular}}
\end{table*}

\begin{table*}
\caption{\textbf{Annotation accuracy of attributions.}}
\centering
\resizebox{0.7\linewidth}{!}{
\begin{tabular}{l l c}
\toprule
Attribution & Tag & Accuracy (\%) \\
\midrule
Lane Direction & Unidirectional / Bidirectional & 99.8 \\
Lane Type &  Solid / Dotted / Fishbone & 99.3\\
Lane Property & General / Stay / Tide / Bus / Three Color & 99.6\\
Lane Flag & Single / Double / Triple & 99.3\\
Lane Width & Normal / Wide & 99.8\\
Curb Type & Ground Side / Road Side / Guardrail & 91.6 \\
Arrow Direction  & Straight / Turn Off / Merge Right / No Turn Left / ... & 98.1\\
Lane Sign Type &  Bike Lane / Bus Lane  & 100.0 \\
Marking Type & Diamond Marking / Inverted Triangle Marking & 92.6\\
\bottomrule
\end{tabular}}
\label{tab:attribution}
\end{table*}

\section{Conclusion and Discussion}
\label{sec:conclusion}
We present VMA, a vectorized map annotation framework based on the unified map representation and the divide-and-conquer annotation scheme.
VMA is highly automatic and extensible, requiring negligible human effort and flexible in terms of spatial scale and element type.
VMA can significantly improve the map generation efficiency and require little human effort, showing great industrial application value.  We find that the quality of scene reconstruction  affects the annotation quality a lot.
By introducing more sensor information (like camera and RADAR) for scene reconstruction, VMA  can be further enhanced. 
And the divide-and-conquer scheme of VMA can also be extended to large-scale lane graph construction based on LaneGAP~\cite{lanegap}.
We leave these as future work.

\bibliographystyle{IEEEtran}
\bibliography{main.bib}

\newpage

\vfill

\end{document}